\documentclass[10pt,twocolumn,letterpaper]{article}

\usepackage[pagenumbers]{wacv} 

\usepackage{graphicx}
\usepackage{amsmath}
\usepackage{amssymb}
\usepackage{booktabs}
\usepackage{lipsum} 
\usepackage{tabularx}

\usepackage[pagebackref,breaklinks,colorlinks]{hyperref}

\usepackage[capitalize]{cleveref}
\crefname{section}{Sec.}{Secs.}
\Crefname{section}{Section}{Sections}
\Crefname{table}{Table}{Tables}
\crefname{table}{Tab.}{Tabs.}

\def\wacvPaperID{1789} 
\def\confName{WACV}
\def\confYear{2025}

\begin{document}

\title{CISOL: An Open and Extensible Dataset for Table Structure Recognition\\ 
in the Construction Industry}

\author{
    David Tschirschwitz \quad Volker Rodehorst\\ \\
    Bauhaus-Universität Weimar, Germany\\
    {\tt\small david.tschirschwitz@uni-weimar.de} 
}
\maketitle

\begin{abstract}
    Reproducibility and replicability are critical pillars of empirical research, particularly in machine learning, where they depend not only on the availability of models, but also on the datasets used to train and evaluate those models. In this paper, we introduce the Construction Industry Steel Ordering List (CISOL) dataset, which was developed with a focus on transparency to ensure reproducibility, replicability, and extensibility. CISOL provides a valuable new research resource and highlights the importance of having diverse datasets, even in niche application domains such as table extraction in civil engineering.

    CISOL is unique in that it contains real-world civil engineering documents from industry, making it a distinctive contribution to the field. The dataset contains more than 120,000 annotated instances in over 800 document images, positioning it as a medium-sized dataset that provides a robust foundation for Table Structure Recognition (TSR) and Table Detection (TD) tasks.
    
    Benchmarking results show that CISOL achieves 67.22 mAP@0.5:0.95:0.05 using the YOLOv8 model, outperforming the TSR-specific TATR model. This highlights the effectiveness of CISOL as a benchmark for advancing TSR, especially in specialized domains.
\end{abstract}

\begin{figure}[ht]
\begin{center}
    \includegraphics[width=\linewidth]{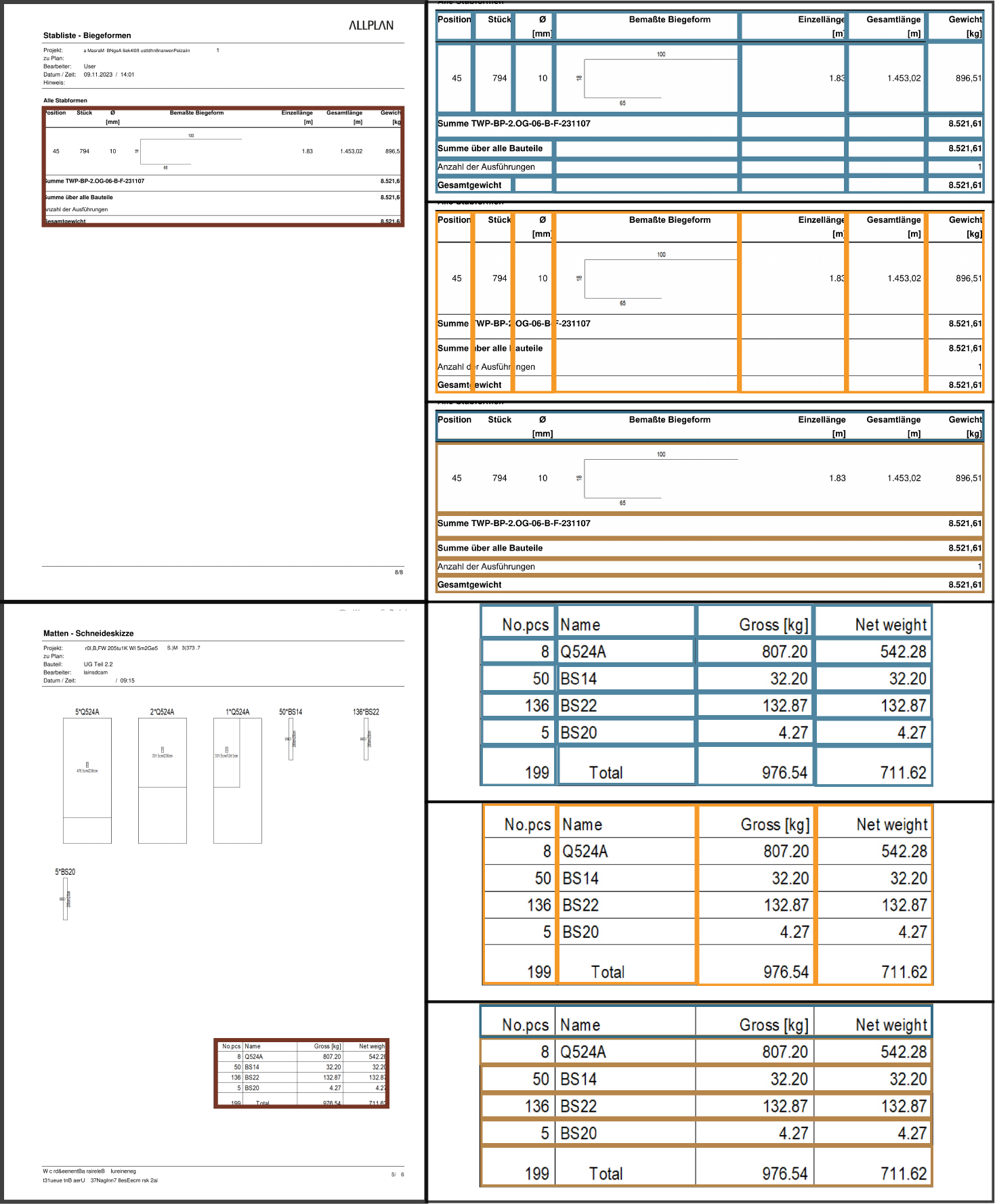}
    \caption{Ground truth annotations of the CISOL dataset on two example document images. The left side shows the table detection task (brown). The right side shows for each of the three images from top to bottom 1) spanning cells (light blue), 2) columns (orange) and 3) rows (beige) and headers (navy) locations. The variety in the table structure and distracting elements such as technical drawings of steel bending techniques show the challenges of the CISOL dataset.}
    \label{fig:teaser_graphic}
\end{center}
\end{figure}

\section{Introduction}
\label{sec:intro}

\begin{figure*}[ht]
\begin{center}
    \includegraphics[width=\linewidth]{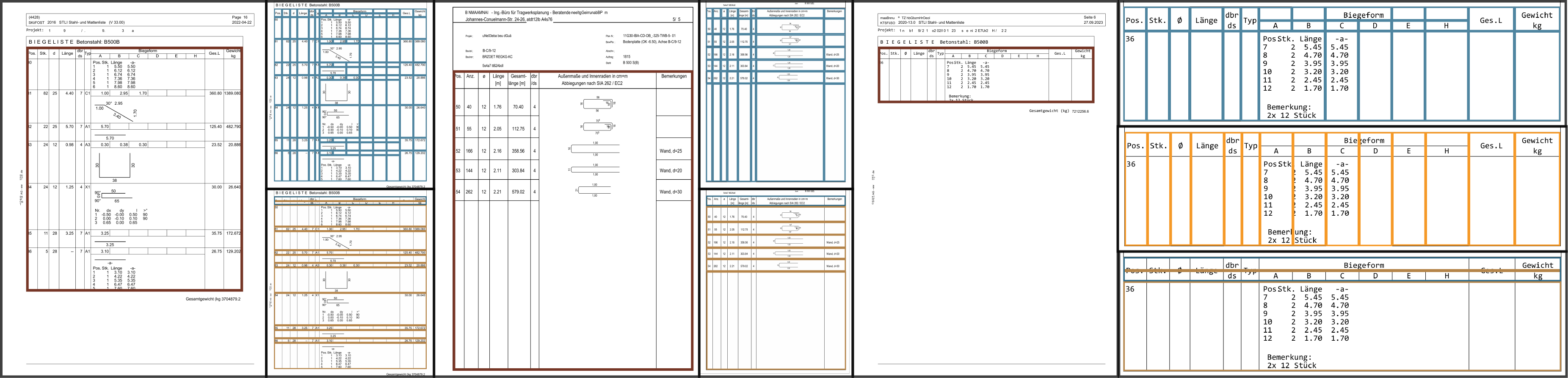}
    \caption{Further visualization of different types of layouts, using the same color coding as Figure \ref{fig:teaser_graphic}. The example on the left shows embedded tables within the actual table, which are considered to be particularly difficult examples. Embedded tables are not annotated in the CISOL dataset and are a prime candidate when extending the dataset. The middle image shows the large variations in the extent to which spanning cells can extend, while the right image shows a case where some spanning cells extend over embedded tables.}
    \label{fig:more_visualizations}
\end{center}
\end{figure*}

Over the past decade, advancements in digitization in document analysis and recognition have been primarily driven by deep learning techniques \cite{liu2018}. As these data-driven methods rely heavily on quality training and evaluation datasets, it is crucial to have a variety of publicly available datasets that allow for comprehensive comparisons across different domains and sub-domains. This is particularly important given the limited generalizability of even the most advanced deep learning models \cite{deeplearningai2021}. Despite the clear importance of the data on which these systems are trained and evaluated, there is a lack of standardization in the creation of datasets. While model development is rigorously pursued to achieve state-of-the-art (SOTA) results, dataset creation often receives less attention, leading to inconsistencies and errors \cite{paullada2021}. Common issues in dataset creation include:
\begin{itemize}
    \item \textbf{Representational harms}: Datasets may be biased towards certain ethnic groups, as observed in some face recognition datasets \cite{paullada2021}.
    \item \textbf{Annotation inconsistency}: Variations in labeling conventions are often ignored or overlooked, with labels being accepted as the "gold standard" without adequate scrutiny \cite{plank2022}.
    \item \textbf{Lack of documentation}: Data collection processes are often opaque and proper documentation is not maintained, hindering the reproducibility and replicability of dataset creation processes \cite{paullada2021, plank2022}.
    \item \textbf{Other issues}: Inadequate licensing for research or commercial use, questionable labor conditions, and poor reusability of data are also prevalent.
\end{itemize}

In this work, we aim to avoid the aforementioned pitfalls by providing a transparent dataset creation process that follows the best practices recommended by Hutchinson \etal \cite{hutchinson2021}, Gebru \etal \cite{gebru2021}, and Paullada \etal \cite{paullada2021}. Our approach is based on the data development lifecycle outlined by Hutchinson \etal \cite{hutchinson2021}, which divides the process into (1) requirements analysis, (2) design, and (3) implementation, while our transparent and extensible approach also supports (4) testing and (5) maintenance of the dataset. The documentation of our dataset is based on the concept of a datasheet as described by Gebru \etal \cite{gebru2021}, and focuses on several core principles:
\begin{itemize}
    \item Replicability and reproducibility.
    \item FAIR principles \cite{wilkinson2016}: Findability, Accessibility, Interoperability, and Reusability.
    \item Extensibility and transparency to ensure maintainability.
    \item Analysis of annotation consistency and the impact of inconsistencies on learning.
    \item Comparability of models by providing a stable and long-lasting evaluation server with unknown test data.
\end{itemize}

\begin{table*}[]
    \newcolumntype{L}{>{\raggedright\arraybackslash}X}
    \begin{tabularx}{\textwidth}{LLLLLll}
    Dataset Name           & Domain                                                               & Image Origin                       & Boundary Type & Evaluation Metric                    & Meta-Data \\
    \toprule
    ICDAR 2013 (Corrected) \cite{gobel2013, smock2023} & Governmental documents                                               & Born-digital documents             & CL            & Precision, recall, F1 score             & no        \\ \hline
    ICDAR 2019 Track-B2 \cite{gao2019a}    & Modern and archival documents                                        & Scanned documents and born-digital & CCL for modern, CL for historic          & Precision, recall, F1 score             & no        \\ \hline
    UNLV \cite{shahab2010}  & Magazines, newspapers, business letters, annual reports              & Scanned documents                  & CL            & Custom / other                                     & no        \\ \hline
    UW3 \cite{shahab2010}   & Books, magazines                                                     & Skew-corrected scanned documents   & CL            & Custom / other                                     & no        \\ \hline
    TabRecSet \cite{yang2023} & Natural images & Search engine crawled photos of tables   & CL & TEDS, precision, average precision &
    no \\ \hline
    TUCD \cite{raja2022} & Business annual reports                                              & Born-digital documents             & CL            & Precision, recall, F1 score             & no        \\ \hline
    WTW \cite{long2021}  & Natural images, Archival document images and printed document images & Photographed or scanned documents  & CL            & Precsion, recall, F1 score, TEDS        & no        \\ \bottomrule
    CISOL (ours)                 & Construction                                                         & Scanned company data               & CL            & COCO mAP                                       & yes      
    \end{tabularx}
    \caption{Qualitative difference of comparable human-annotated TSR datasets. It differentiates between the domains, image origin, the type of cell boundary, the evaluation metric used for the visual identification of the structural elements and the provision of meta-data.}
    \label{tab:dataset_comparison}
\end{table*}

The CISOL dataset presented in this paper focuses on table extraction within the civil engineering domain. As categorized by Smock \etal \cite{smock2022}, our dataset supports table detection and table structure recognition, but does not extend to table functional analysis as illustrated in Figure \ref{fig:teaser_graphic} and \ref{fig:more_visualizations}. The primary tasks for models trained on this dataset are to detect tables, identify columns, rows and cells, including spanning cells. In addition, we provide headers that can be used for functional analysis. Table extraction can be further categorized based on the type of structure detected: (a) physical structure and (b) logical structure. The community is divided on which structure is preferable for analysis. Physical structure detection offers better extensibility \cite{huang2023} and is advantageous for downstream tasks such as text information extraction or table question answering, although it requires post-processing to derive a usable table structure. Conversely, the logical structure provides a more accurate representation of the table organization \cite{lysak2023}, but provides less visual information and requires post-processing to obtain bounding box locations \cite{siddiqui2019}. Given the principles outlined above, we have chosen to focus on the physical structure for this dataset.

Tables provide a valuable format for encoding and structuring information \cite{lysak2023}, but present challenges due to (1) intra-class variability \cite{schreiber2017} and diversity \cite{hashmi2021}, (2) the presence of embedded tables, as found in our dataset, (3) complex table structures \cite{schreiber2017}, and (4) the variety of table styles and formats \cite{lysak2023}, including the unpredictable use of rule lines \cite{schreiber2017}. Consequently, table extraction remains a challenging and current problem within the document analysis and recognition community \cite{lysak2023}.

The core contribution of this paper is the introduction of a new dataset (CISOL) for table extraction, which is characterized by a combination of the following features:
\begin{enumerate}
    \item Domain specificity to civil engineering, a domain for which no such dataset currently exists.
    \item Anonymized real-world industry data, typically inaccessible, provided with appropriate licensing for research use.
    \item An open guideline and an extensible, transparent dataset creation process.
    \item Easily accessible data and an open, long-term evaluation server with unpublished test annotations for benchmarking purposes.
\end{enumerate}

\section{Related Work}

Table extraction datasets can be categorized into two types based on their labeling approach: (1) human-labeled data and (2) synthetic-labeled data. Synthetic-labeled datasets, such as PubTabNet \cite{zhong2020}, FinTabNet \cite{zheng2021}, and others \cite{li2020e, deng2019, smock2022}, use heuristics to extract the logical or physical table structure from formats such as HTML or LaTeX files. This approach requires that all documents are born-digital, i.e. they have never been printed and scanned. While datasets created by synthetic labeling strategies are often large, they are limited to data that is already automatically parsable. As a result, their usefulness for digitizing documents depends on a model's ability to transfer between domains, or for training foundational models that can be fine-tuned on data relevant to specific use cases. We do not compare our dataset to this type of data.

\begin{figure*}[ht]
\begin{center}
    \includegraphics[width=\linewidth]{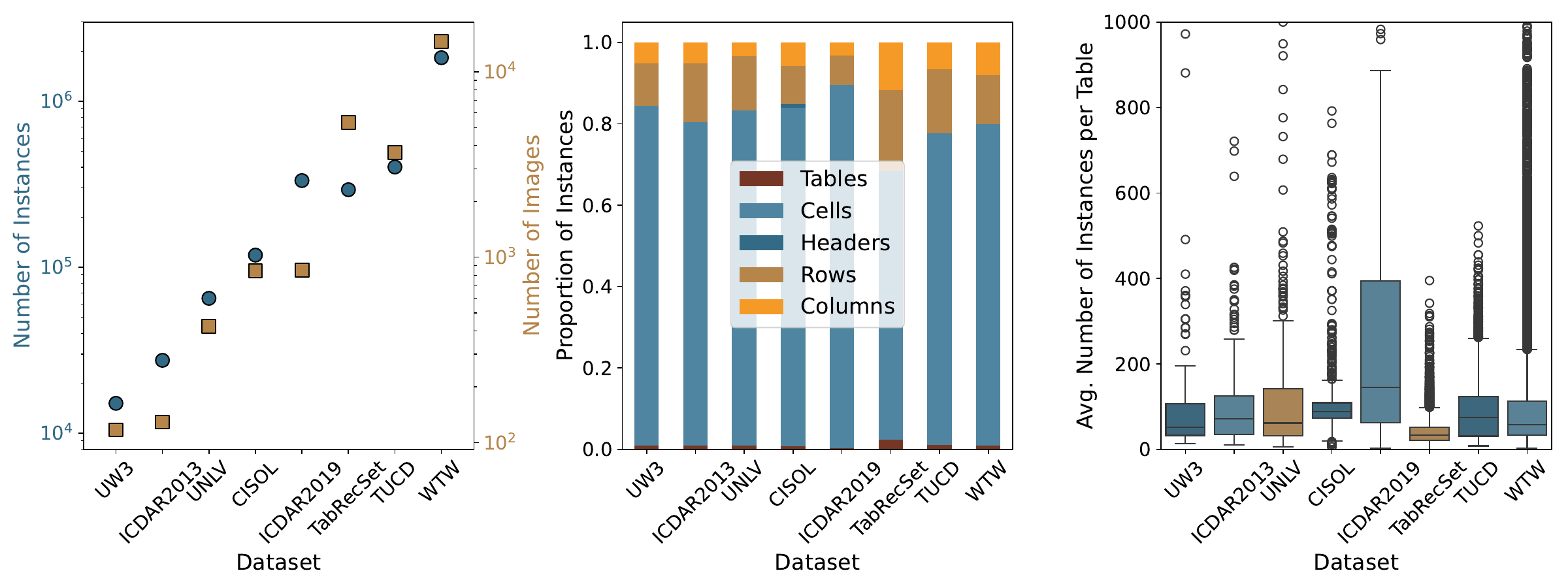}
    \caption{On the left, the number of instances and images for each of the datasets are compared on a logarithmic scale, showing that CISOL is a medium sized dataset. The middle plot shows the class ratio for the five different datasets, a lower number of rows would indicate wider tables, while a lower number of columns would indicate narrower tables. The number of cells also gives an idea of the size of the tables. In the right plot, the density of annotations per document image can be analyzed, providing information on the variation within the dataset. For CISOL, while a large number of tables have a similar number of instances, there are a large number of outliers at the lower and upper ends. Some outliers exist beyond 1000 instances, but are not plotted to keep the plot zoomed in.}
    \label{fig:dataset_comparison}
\end{center}
\end{figure*}

Human-labeled data, on the other hand, are not restricted in terms of the type of data that can be annotated. Most human-labeled datasets focus on scanned documents. These types of datasets are more expensive and labor-intensive to create, presenting a more challenging task due to the smaller amount of training data and the often more varied documents compared to synthetic-labeled data. We compare the presented dataset with existing research resources, but exclude datasets that focus on only one of the two subtasks, such as Marmot \cite{fang2012}, RVL-CDIP \cite{harley2015}, and IIIT-AR-13K \cite{mondal2020}, which are limited to table detection. We also exclude ICDAR-2017-POD \cite{gao2017} and TabStructDB \cite{siddiqui2019}, as they lack table detection and do not include spanning cell information. Finally, we excluded the IFLYTAB \cite{zhang2024} and FinTab \cite{li2021gfte} datasets as they were not accessible at this time. Datasets that encompass both subtasks and extend beyond them are included, but are only compared on the two subtasks.

The CISOL dataset is then compared to the seven remaining relevant datasets in Table \ref{tab:dataset_comparison} based on their qualitative differences, such as domain, image origin, and other factors. Additionally, Figure \ref{fig:dataset_comparison} provides a quantitative analysis of the number of instances and images, class distribution, and annotation density. For the WTW \cite{long2021} dataset the revised test data was used and columns, rows and tables were inferred from the cells and for the ICDAR 2013 \cite{gobel2013} dataset we used the corrected version by Smock \etal \cite{smock2023}.

Through this analysis, CISOL can be considered a medium-sized dataset, characterized by generally larger tables, as indicated by the annotation density and class distribution. Beyond the introduction of a new domain in the selection of table extraction datasets, the main advantage of CISOL lies in its extensibility and the availability of metadata, which is not limited to the original dataset creators. As the annotation pipeline is simple and straightforward to reproduce, it is possible to replicate the annotation process or extend it to other sub-domains. This may eliminate the need to create a new annotation guideline, instead allowing the existing one to be extended with newly acquired knowledge.

The selection of the evaluation metric was guided by the visually focused task of detecting table elements, so we adopted the evaluation metric used for the PubTabNet \cite{zhong2020} Challenge Track A held at ICDAR 2021, which uses the classical Mean Average Precision @ 0.5:0.05:0.95 metric, as known from the COCO dataset \cite{CoCo}.

\section{Dataset}

The Construction Industry Steel Ordering List (CISOL) dataset adds data from industry sources the pool of available table extraction datasets, which are largely derived from publicly available government or academic documents. The following three sub-sections explain the design decision, the creation process and the comparative statistics.

\begin{figure*}[ht]
\begin{center}
    \includegraphics[width=0.9\linewidth]{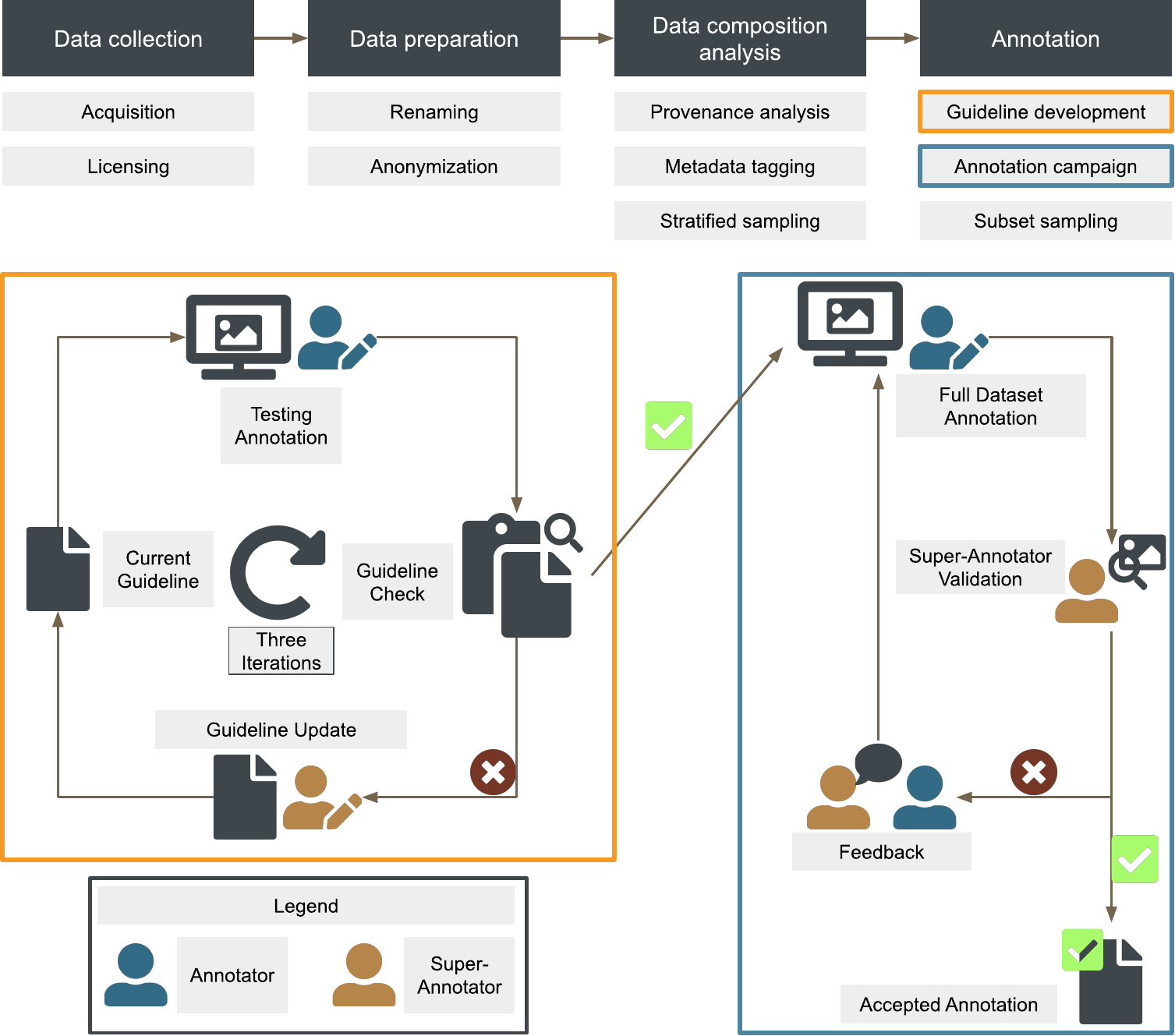}
    \caption{Annotation pipeline of CISOL following four main stages, including two iterative approaches to guideline creation and the annotation campaign. The use of publicly available tools \cite{CVAT_ai_Corporation_Computer_Vision_Annotation_2023} and a straightforward, simple annotation pipeline makes the process easy to replicate and reproduce.}
    \label{fig:annotation_pipeline}
\end{center}
\end{figure*}

\subsection{Design Decisions}

The dataset is designed to effectively address the problem of Table Structure Recognition (TSR) by directly identifying table components visually via learned techniques and then using simple heuristics to construct the table structure, minimizing the need for speculative inference within these heuristics. This design choice is particularly important when dealing with complex elements such as headers and spanning cells. By accurately identifying these elements, the dataset allows the logical structure of the table to be inferred directly from its physical arrangement using simple heuristics. This approach not only simplifies the extraction of logical structures, but also provides optimal support for downstream tasks such as table content recognition and embedded table recognition  \cite{huang2023}.

However, the scope of the dataset is constrained by the specific company data available, which in this case focuses on the building sector. The dataset consists primarily of tables detailing various aspects of steel orders, including physical dimensions such as length, weight, and diameter, as well as technical drawings illustrating steel bending techniques. This focused design ensures that the dataset is well-suited to address the TSR problem within its specific domain, facilitating both the direct identification of table components and the logical structure extraction via heuristics.

An important aspect of all design decisions was to allow for extensibility of the dataset, so all types of tools, the creation pipeline and the guideline must be publicly available.

\subsection{Creation Process}

\begin{figure*}[ht]
\begin{center}
    \includegraphics[width=\linewidth]{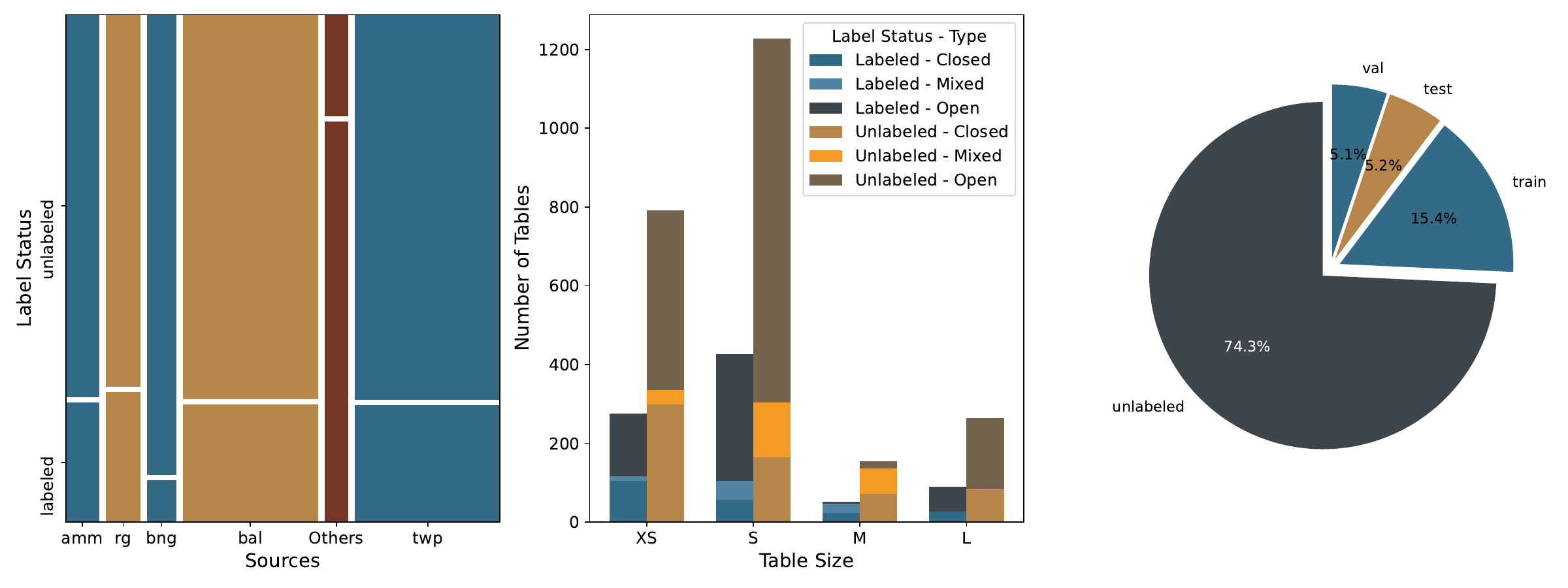}
    \caption{Overview of data origin (left), selection for annotation (center), and dataset splits (right). The left chart shows data sources anonymized by company abbreviation, with 'Others' representing a mix of smaller contributors. Annotations have been prioritized for rare instances, with the remaining samples randomly selected. The middle chart compares the distribution of metadata tags between annotated and unannotated data, illustrating the balance of types and sizes. Finally, the right chart shows the split of annotated data into training, validation, and test sets.}
    \label{fig:cisol_analysis}
\end{center}
\end{figure*}

The creation of CISOL follows four main steps: (1) data collection, (2) data preparation, (3) data composition analysis, and (4) annotation, as outlined in Figure \ref{fig:annotation_pipeline}.

During the \textbf{data collection} phase, data were gathered from 24 unique reinforced concrete projects, contributed by 10 different structural engineering firms, with creation dates ranging from 2015 to 2023. This resulted in a total of 3,288 document images. These documents, all in German, reflect the geographical context of construction projects across Germany. The data is licensed under the Creative Commons Attribution 4.0 International license to support full research use, subject to proper anonymization during the preparation phase.

In the \textbf{data preparation} phase, the data undergoes several processing steps, including renaming and organizing files to facilitate automated processing, such as converting PDF files to images. An essential aspect of this phase is the anonymization process, which ensures compliance with the licensing terms. Anonymization is performed in two stages: an automated process and a manual review. The automated anonymization identifies predefined text patterns, such as company names, locations, and personal data, and replaces them with text of similar length and capitalization, ensuring that the overall document layout remains largely unchanged and the original files are preserved as closely as possible. This is followed by a manual review to identify any remaining identifiable information, such as company logos or other details not captured by the automated process. Both the automated and manual anonymization steps are documented in the metadata file associated with the dataset. During this process, eight document images were removed because they contained only project-specific information, such as cover pages, without relevant tables.

In the third step, the \textbf{data composition analysis} is conducted to generate metadata that helps to understand the provenance, table type, and size of the different document images. This metadata is crucial to ensure that the annotated data accurately represents the targeted problem, or at least provides information on the type of problem for which the dataset can be used. As the data was collected based on availability, this analysis helps to select the most representative samples for the annotation process. Figure \ref{fig:cisol_analysis} illustrates the different stages of this process. While some provenance metadata is available from the data acquisition phase, additional manual steps are taken to tag the dataset. Two tags are assigned to each document image:

\begin{enumerate} 
    \item \textbf{Table size}: Categorized as XS for tables with less than 50 cells, S for tables with 50-100 cells, M for tables with 100-200 cells, and L for tables with more than 200 cells. These categories are approximations based on visual inspection and may vary depending on factors such as the number of spanning cells. 
    \item \textbf{Table type}: Classified as open, closed, or mixed, based on the presence and arrangement of separator lines. Closed tables have both vertical and horizontal dividers between cells, mixed tables have some separator lines, and open tables have none. 
\end{enumerate}

These additional tags allow a methodological categorization of the data, which serves two main purposes: (1) to facilitate stratified sampling for annotation and the splitting of the data into training, validation, and test sets, ensuring an equal representation of different types, sizes, and origins; and (2) to optimize the workload distribution among the annotators. By applying these tools, the limited annotation budget can be used efficiently to represent the challenges of different table extraction scenarios.

Stratified sampling was based on the origin of the data, specifically the company that created it, with a preference for annotating data from companies with fewer than 100 documents. This approach ensures that a wider range of company data is included. However, in order to preserve the natural distribution of table sizes and types, stratified sampling was not applied to these criteria. The random sampling process was repeated several times to achieve a sample with a distribution similar to the full dataset, with respect to size and type.

The core step of the dataset creation process is the human \textbf{annotation} of the selected subset. This process is divided into two phases: the creation of the annotation guideline and the subsequent annotation campaign, as shown in Figure \ref{fig:annotation_pipeline}. Both phases are overseen by a super-annotator, who guides the annotation process and is responsible for adapting the guideline to ensure that a learnable problem is embedded in the process. This means defining clear and distinct classes, while thoroughly covering edge cases and leaving little room for annotators to establish their own labeling conventions.

The Computer Vision Annotation Tool (CVAT) \cite{CVAT_ai_Corporation_Computer_Vision_Annotation_2023} is used for both guideline creation and annotation. The guideline creation begins with an initial draft, which is then given to a test annotator who annotates a small subset of the data. If the test annotations reveal any ambiguities in the guideline—where the interpretation was too open or the annotator's work deviated from the super-annotator's expectations—the guideline is revised. This iterative process continues until the guideline is clear and consistent, ensuring that subjective labeling conventions are minimized. 

Once the guideline is finalized, the annotation rollout begins with a kickoff meeting led by the super-annotator, who briefs all annotators on the process. Each annotator is then assigned equal work packages to complete. Using the functions provided in CVAT \cite{CVAT_ai_Corporation_Computer_Vision_Annotation_2023}, annotators submit their completed jobs for validation by the super-annotator. If any issues are identified, the job is returned to the annotator with comments for revision. If the annotations are satisfactory, they are added to the dataset. During this process, additional edge cases may be added to the guideline as they arise.

The annotators involved in this process had undergraduate academic qualifications. However, we believe that no special qualifications are necessary to create these annotations, and that the task can be performed by laypersons, as the guideline provides the necessary knowledge. This simple, reproducible process allows for easy extension of the existing dataset. 

Once the dataset was finalized, it was split into training, validation and test subsets using the three tags of origin, size and type to balance the dataset between the subsets. The dataset and the annotation guideline can be found at \url{https://doi.org/10.5281/zenodo.10829550} with COCO-JSON (\url{https://cocodataset.org/#format-data}) used as the annotation format.

\subsection{Annotation Consistency, Convergence Threshold and Benchmarking}

\begin{figure}[h]
\begin{center}
    \includegraphics[width=\linewidth]{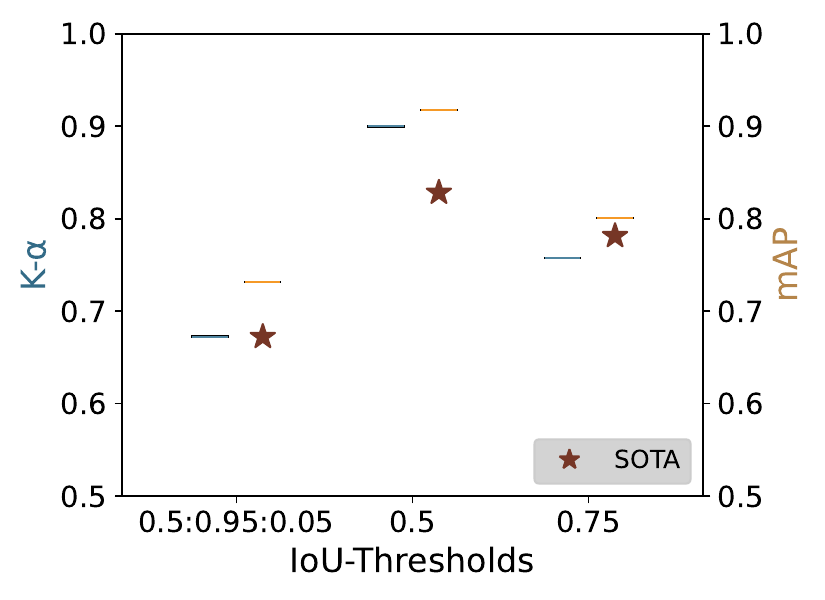}
    \caption{Annotation consistency values ($K-\alpha$) and convergence threshold (mAP) for different IoU thresholds compared to the SOTA results. In blue are the $K-\alpha$ values and in beige the convergence threshold, followed by the SOTA star in orange.}
    \label{fig:benchmark}
\end{center}
\end{figure}

To evaluate the quality of annotations, 20 randomly selected samples were re-annotated by all four annotators. These annotations were then compared for consistency using a measure known as Krippendorff's $\alpha$ ($K-\alpha$), as presented by Tschirschwitz \etal \cite{tschirschwitz2022}. The results are plotted in Figure \ref{fig:benchmark}, along with the convergence threshold and benchmarking results. According to the interpretation of the $K-\alpha$ values, scores above 0.667 indicate moderate agreement—this is the case for the thresholds $K-\alpha$@0.5:0.95:0.05 and $K-\alpha$@0.75. Values above 0.8 indicate reliable and strong agreement among raters, which is achieved at the $K-\alpha$@0.5 threshold. Overall, this suggests that the dataset has moderate to high agreement, indicating a well-designed annotation guideline and process.

Based on the $K-\alpha$ value, it is possible to determine the label convergence threshold. The label convergence threshold represents the theoretical upper limit that a model should not be able to exceed due to inconsistencies in the labeling conventions used by annotators. A simple formula is used to derive the mAP values from the $K-\alpha$ values \cite{tschirschwitz2025a}. Using the mAP values, comparability with model performance is possible. The SOTA results are plotted below the convergence thresholds. These results indicate that the problem posed by CISOL is not yet fully solved, leaving room for further model development and fine-tuning. 

As an extension to the current pool of benchmarking datasets, CISOL provides an opportunity to test and improve the generalizability of new methodological approaches. The current SOTA results were obtained using YOLOv8 \cite{yolov8}, rather than a TSR-specific model such as TATR \cite{smock2022}, which was also trained on our dataset but did not perform as well. This is consistent with the findings of Ajayi \etal \cite{ajayi2023}, who found a lack of reproducibility with respect to new datasets in TSR specific models. Since this paper focuses primarily on presenting a new dataset rather than developing a new model, the results can be considered satisfactory, as indicated by their proximity to the theoretical upper bound.

The exact values of the current best performing model on the CISOL dataset can be found on the evaluation server: \url{https://eval.ai/web/challenges/challenge-page/2257}.

\section{Conclusion}

In summary, the CISOL dataset introduces a novel resource for table structure recognition that specifically addresses the unique challenges posed by real-world documents in the civil engineering domain. Unlike previous datasets, CISOL provides anonymized, licensed, real-world company data, providing a valuable benchmark for models focused on table detection and structure recognition. The extensibility, transparency, and open guidelines of the dataset allow researchers to easily extend and adapt the dataset to new subdomains, making it a versatile tool for ongoing research.

Through rigorous annotation processes and benchmarking, CISOL demonstrates moderate to high agreement among annotators, ensuring the reliability of the dataset. It serves as an essential resource for training models that can be applied to downstream tasks, such as table content recognition or the handling of complex table layouts, including embedded tables. By comparing the convergence threshold with high-performing object detection models and TSR-specific models, we have ensured that CISOL poses a significant challenge to current research, making it highly relevant for future model development. In addition, CISOL can be added to the selection of datasets used for benchmarking table extraction, expanding the range of domains and improving the generalizability of models developed for this task.

Looking ahead, there are several opportunities to improve and extend the CISOL dataset. Future work could involve the inclusion of embedded tables, which would increase the depth of the dataset and address more complex table structures. Additionally, extending the dataset to include logical table structures or even cell content would provide a richer and more detailed resource for training models that require deeper semantic understanding. Expanding the breadth of the dataset to include new domains, potentially within civil engineering or other industries, would make CISOL a more comprehensive benchmark capable of supporting a wider range of table extraction tasks. These enhancements would strengthen CISOL's role as a resource for advances in table structure recognition.

{\small
\bibliographystyle{ieee_fullname}
\bibliography{main}

\begin{thebibliography}{10}\itemsep=-1pt

\bibitem{ajayi2023}
Kehinde Ajayi, Muntabir~Hasan Choudhury, Sarah Rajtmajer, and Jian Wu.
\newblock A {{Study}} on {{Reproducibility}} and {{Replicability}} of {{Table Structure Recognition Methods}}.
\newblock In {\em Proceedings of the International Conference on Document Analysis and Recognition (ICDAR)}, 2023.

\bibitem{CVAT_ai_Corporation_Computer_Vision_Annotation_2023}
{CVAT.ai Corporation}.
\newblock {Computer Vision Annotation Tool (CVAT)}, Nov. 2023.

\bibitem{deeplearningai2021}
{DeepLearningAI}.
\newblock A {Chat} with {Andrew} on {MLOps}: {From} {Model}-centric to {Data}-centric {AI}, 2021.

\bibitem{deng2019}
Yuntian Deng, David Rosenberg, and Gideon Mann.
\newblock Challenges in {{End-to-End Neural Scientific Table Recognition}}.
\newblock In {\em Proceedings of the International Conference on Document Analysis and Recognition (ICDAR)}, 2019.

\bibitem{fang2012}
Jing Fang, Xin Tao, Zhi Tang, Ruiheng Qiu, and Ying Liu.
\newblock Dataset, {{Ground-Truth}} and {{Performance Metrics}} for {{Table Detection Evaluation}}.
\newblock In {\em Proceedings of the International Workshop on Document Analysis Systems (DAS)}, 2012.

\bibitem{gao2019a}
Liangcai Gao, Yilun Huang, Hervé Déjean, Jean-Luc Meunier, Qinqin Yan, Yu Fang, Florian Kleber, and Eva Lang.
\newblock {{ICDAR}} 2019 {{Competition}} on {{Table Detection}} and {{Recognition}} ({{cTDaR}}).
\newblock In {\em Proceedings of the International Conference on Document Analysis and Recognition (ICDAR)}, 2019.

\bibitem{gao2017}
Liangcai Gao, Xiaohan Yi, Zhuoren Jiang, Leipeng Hao, and Zhi Tang.
\newblock {{ICDAR2017 Competition}} on {{Page Object Detection}}.
\newblock In {\em Proceedings of the International Conference on Document Analysis and Recognition (ICDAR)}, 2017.

\bibitem{gebru2021}
Timnit Gebru, Jamie Morgenstern, Briana Vecchione, Jennifer~Wortman Vaughan, Hanna Wallach, Hal~Daumé Iii, and Kate Crawford.
\newblock Datasheets for datasets.
\newblock {\em Communications of the ACM (CAMC)}, 2021.

\bibitem{gobel2013}
Max Gobel, Tamir Hassan, Ermelinda Oro, and Giorgio Orsi.
\newblock {{ICDAR}} 2013 {{Table Competition}}.
\newblock In {\em Proceedings of the International Conference on Document Analysis and Recognition (ICDAR)}, 2013.

\bibitem{harley2015}
Adam~W. Harley, Alex Ufkes, and Konstantinos~G. Derpanis.
\newblock Evaluation of deep convolutional nets for document image classification and retrieval.
\newblock In {\em Proceedings of the International Conference on Document Analysis and Recognition (ICDAR)}, 2015.

\bibitem{hashmi2021}
Khurram~Azeem Hashmi, Marcus Liwicki, Didier Stricker, Muhammad~Adnan Afzal, Muhammad~Ahtsham Afzal, and Muhammad~Zeshan Afzal.
\newblock Current {{Status}} and {{Performance Analysis}} of {{Table Recognition}} in {{Document Images With Deep Neural Networks}}.
\newblock {\em IEEE Access}, 2021.

\bibitem{huang2023}
Yongshuai Huang, Ning Lu, Dapeng Chen, Yibo Li, Zecheng Xie, Shenggao Zhu, Liangcai Gao, and Wei Peng.
\newblock Improving {{Table Structure Recognition}} with {{Visual-Alignment Sequential Coordinate Modeling}}.
\newblock In {\em Proceedings of the IEEE/CVF Conference on Computer Vision and Pattern Recognition (CVPR)}, 2023.

\bibitem{hutchinson2021}
Ben Hutchinson, Andrew Smart, Alex Hanna, Emily Denton, Christina Greer, Oddur Kjartansson, Parker Barnes, and Margaret Mitchell.
\newblock Towards {{Accountability}} for {{Machine Learning Datasets}}: {{Practices}} from {{Software Engineering}} and {{Infrastructure}}.
\newblock In {\em Proceedings of the ACM Conference on Fairness, Accountability and Transparency (FAccT)}, 2021.

\bibitem{yolov8}
Glenn Jocher, Ayush Chaurasia, and Jing Qiu.
\newblock {Ultralytics YOLO}, Jan. 2023.

\bibitem{li2020e}
Minghao Li, Lei Cui, Shaohan Huang, Furu Wei, Ming Zhou, and Zhoujun Li.
\newblock {{TableBank}}: {{Table Benchmark}} for {{Image-based Table Detection}} and {{Recognition}}.
\newblock In Nicoletta Calzolari, Frédéric Béchet, Philippe Blache, Khalid Choukri, Christopher Cieri, Thierry Declerck, Sara Goggi, Hitoshi Isahara, Bente Maegaard, Joseph Mariani, Hélène Mazo, Asuncion Moreno, Jan Odijk, and Stelios Piperidis, editors, {\em International Conference on Language Resources and Evaluation (LREC)}, 2020.

\bibitem{li2021gfte}
Yiren Li, Zheng Huang, Junchi Yan, Yi Zhou, Fan Ye, and Xianhui Liu.
\newblock Gfte: graph-based financial table extraction.
\newblock In {\em Proceedings of the Internationl Conference on Pattern Recognition (ICPR) International Workshops and Challenges}, 2021.

\bibitem{CoCo}
Tsung-Yi Lin, Michael Maire, Serge Belongie, James Hays, Pietro Perona, Deva Ramanan, Piotr Doll{\'a}r, and C~Lawrence Zitnick.
\newblock Microsoft {COCO}: {Common} {Objects} in {Context}.
\newblock In {\em Proceedings of the European Conference on Computer Vision (ECCV)}, 2014.

\bibitem{liu2018}
Cheng-Lin Liu, Gernot~A. Fink, Venu Govindaraju, and Lianwen Jin.
\newblock Special issue on deep learning for document analysis and recognition.
\newblock {\em International Journal on Document Analysis and Recognition (IJDAR)}, 2018.

\bibitem{long2021}
Rujiao Long, Wen Wang, Nan Xue, Feiyu Gao, Zhibo Yang, Yongpan Wang, and Gui-Song Xia.
\newblock Parsing {{Table Structures}} in the {{Wild}}.
\newblock In {\em Proceedings of the IEEE/CVF International Conference on Computer Vision (ICCV)}, 2021.

\bibitem{lysak2023}
Maksym Lysak, Ahmed Nassar, Nikolaos Livathinos, Christoph Auer, and Peter Staar.
\newblock Optimized {{Table Tokenization}} for {{Table Structure Recognition}}.
\newblock In {\em Proceedings of the International Conference on Document Analysis and Recognition (ICDAR)}, 2023.

\bibitem{mondal2020}
Ajoy Mondal, Peter Lipps, and C.~V. Jawahar.
\newblock {{IIIT-AR-13K}}: {{A New Dataset}} for {{Graphical Object Detection}} in {{Documents}}.
\newblock In Xiang Bai, Dimosthenis Karatzas, and Daniel Lopresti, editors, {\em Proceedings of the International Workshop on Document Analysis Systems (DAS)}, 2020.

\bibitem{paullada2021}
Amandalynne Paullada, Inioluwa~Deborah Raji, Emily~M. Bender, Emily Denton, and Alex Hanna.
\newblock Data and its (dis)contents: {{A}} survey of dataset development and use in machine learning research.
\newblock {\em Patterns}, 2021.

\bibitem{plank2022}
Barbara Plank.
\newblock The {``}problem{''} of human label variation: On ground truth in data, modeling and evaluation.
\newblock In {\em Proceedings of the Conference on Empirical Methods in Natural Language Processing (EMNLP)}, 2022.

\bibitem{raja2022}
Sachin Raja, Ajoy Mondal, and Jawahar~C V.
\newblock Visual {{Understanding}} of {{Complex Table Structures}} from {{Document Images}}.
\newblock In {\em Proceedings of the IEEE/CVF Winter Conference on Applications of Computer Vision (WACV)}, 2022.

\bibitem{schreiber2017}
Sebastian Schreiber, Stefan Agne, Ivo Wolf, Andreas Dengel, and Sheraz Ahmed.
\newblock {{DeepDeSRT}}: {{Deep Learning}} for {{Detection}} and {{Structure Recognition}} of {{Tables}} in {{Document Images}}.
\newblock In {\em Proceedings of the International Conference on Document Analysis and Recognition (ICDAR)}, 2017.

\bibitem{shahab2010}
Asif Shahab, Faisal Shafait, Thomas Kieninger, and Andreas Dengel.
\newblock An open approach towards the benchmarking of table structure recognition systems.
\newblock In {\em Proceedings of the International Workshop on Document Analysis Systems (DAS)}, 2010.

\bibitem{siddiqui2019}
Shoaib~Ahmed Siddiqui, Imran~Ali Fateh, Syed Tahseen~Raza Rizvi, Andreas Dengel, and Sheraz Ahmed.
\newblock {{DeepTabStR}}: {{Deep Learning}} based {{Table Structure Recognition}}.
\newblock In {\em {{ICDAR}}}, 2019.

\bibitem{smock2022}
Brandon Smock, Rohith Pesala, and Robin Abraham.
\newblock {{PubTables-1M}}: {{Towards}} comprehensive table extraction from unstructured documents.
\newblock In {\em Proceedings of the IEEE/CVF Conference on Computer Vision and Pattern Recognition (CVPR)}, 2022.

\bibitem{smock2023}
Brandon Smock, Rohith Pesala, and Robin Abraham.
\newblock Aligning benchmark datasets for table structure recognition.
\newblock In {\em Proceedings of the International Conference on Document Analysis and Recognition (ICDAR)}, 2023.

\bibitem{tschirschwitz2022}
David Tschirschwitz, Franziska Klemstein, Benno Stein, and Volker Rodehorst.
\newblock A {Dataset} for {Analyzing} {Complex} {Document} {Layouts} in the {Digital} {Humanities} and its {Evaluation} with {Krippendorff} ’s {Alpha}.
\newblock In {\em Proceedings of the German Conference on Pattern Recognition (GCPR)}, 2022.

\bibitem{tschirschwitz2025a}
David Tschirschwitz and Volker Rodehorst.
\newblock Label convergence: Defining the upper performance bound in object recognition through contradictory annotations.
\newblock In {\em Proceedings of the IEEE/CVF Winter Conference on Applications of Computer Vision (WACV)}, 2025.

\bibitem{wilkinson2016}
Mark~D. Wilkinson and \etal.
\newblock The {{FAIR Guiding Principles}} for scientific data management and stewardship.
\newblock {\em Scientific Data}, 2016.

\bibitem{yang2023}
Fan Yang, Lei Hu, Xinwu Liu, Shuangping Huang, and Zhenghui Gu.
\newblock A large-scale dataset for end-to-end table recognition in the wild.
\newblock {\em Scientific Data}, 2023.

\bibitem{zhang2024}
Zhenrong Zhang, Pengfei Hu, Jiefeng Ma, Jun Du, Jianshu Zhang, Baocai Yin, Bing Yin, and Cong Liu.
\newblock Semv2: Table separation line detection based on instance segmentation.
\newblock {\em Pattern Recognition}, 2024.

\bibitem{zheng2021}
Xinyi Zheng, Douglas Burdick, Lucian Popa, Xu Zhong, and Nancy Xin~Ru Wang.
\newblock Global {{Table Extractor}} ({{GTE}}): {{A Framework}} for {{Joint Table Identification}} and {{Cell Structure Recognition Using Visual Context}}.
\newblock In {\em Proceedings of the IEEE/CVF Winter Conference on Applications of Computer Vision (WACV)}, 2021.

\bibitem{zhong2020}
Xu Zhong, Elaheh ShafieiBavani, and Antonio~Jimeno Yepes.
\newblock Image-based table recognition: Data, model, and evaluation.
\newblock In {\em Proceedings of the European Conference on Computer Vision (ECCV)}, 2020.

\end{thebibliography}
}



\end{document}